\documentclass{article}

\usepackage{PRIMEarxiv}

\usepackage[utf8]{inputenc} 
\usepackage[T1]{fontenc}    
\usepackage{hyperref}       
\usepackage{url}            
\usepackage{booktabs}       
\usepackage{amsfonts}       
\usepackage{nicefrac}       
\usepackage{microtype}      
\usepackage{lipsum}
\usepackage{fancyhdr}       
\usepackage{graphicx}       
\graphicspath{{media/}}     

\usepackage{multirow}

\title{Intelligent Interface: Enhancing Lecture Engagement with Didactic Activity Summaries
}

\author{
Anna Wróblewska, Marcel Witas, Kinga Frańczak, Arkadiusz Kniaź\\
Faculty of Mathematics and Information Sciences,
Warsaw University of Technology\\
 Warsaw, Poland\\
\texttt{anna.wroblewska1@pw.edu.pl}\\
\And Cheong Siew Ann\\
School of Physical and Mathematical Sciences, Nanyang Technological University, Singapore \\
\And Seng Chee Tan \\
National Institute of Education, Nanyang Technological Institute, Singapore\\
\And Janusz Hołyst\\
Faculty of Physics, Warsaw University of Technology,  Warsaw, Poland \\
\And Marcin Paprzycki\\
Systems Research Institute Polish Academy of Sciences, Warsaw, Poland\\
}

\begin{document}
\maketitle

\begin{abstract}
Recently, multiple applications of machine learning have been introduced. They include various possibilities arising when image analysis methods are applied to, broadly understood, video streams. In this context, a novel tool, developed for academic educators to enhance the teaching process by automating, summarizing, and offering prompt feedback on conducting lectures, has been developed. The implemented prototype utilizes machine learning-based techniques to recognise selected didactic and behavioural teachers' features within lecture video recordings. 
Specifically, users (teachers) can upload their lecture videos, which are preprocessed and analysed using machine learning models. Next, users can view summaries of recognized didactic features through interactive charts and tables. Additionally, stored ML-based prediction results support comparisons between lectures based on their didactic content. In the developed application text-based models trained on lecture transcriptions, with enhancements to the transcription quality, by adopting an automatic speech recognition solution are applied. Furthermore, the system offers flexibility for (future) integration of new/additional machine-learning models and software modules for image and video analysis.

\end{abstract}

\keywords{Artificial intelligence in education 
\and Lecture evaluation 
\and Natural language processing \and Machine learning. 
}

\section{Introduction}
It is easy to notice that the majority of newly proposed solutions, adopting artificial intelligence (AI) in higher education, focus on supporting students. This involves student learning~\cite{kuvcak2018machine,unescogenai}, measuring student performance~\cite{anand2018recursive}, or assessing available learning sources~\cite{zhu2015machine}. Moreover, since approximately 2022, great interest in utilising and evaluating use of generative AI in education, for learning and teaching~\cite{holmes2023artificial} has emerged. 

A much less frequently addressed subject is the support that could be provided for the teachers, by observing them while they are conducting lessons, and providing personalized feedback. 
Only very few AI-based solutions have been developed to address this need, e.g. tools for observing students' behaviour and cooperation in teamwork during class~\cite{segal2017keeping}, as well as other activities~\cite{broussard2021interface,tarantini2022reflective}. Another type of support -- related to preparing lessons -- includes AI-based tutors developed to support teachers in schools by directing them towards achieving educational goals~\cite{AIcoach}. One more recent tool, is described in~\cite{guo2021ai} and is closely related to the approach proposed in this contribution. Here, the work describes a system that includes teaching evaluation indicators that combines traditional assessment scales with new values derived from the computer vision, audio speech recognition, and machine learning. while quite interesting, it provides only a comprehensive behavioural classification of the teacher, e.g. teacher's style (i.e. humorous, passionate, solemn), or teacher's media usage (i.e. multimedia vs blackboard). Research reported here proposes to take the teacher support, using already available AI-based methods, a step further. Specifically, the main goal is to improve the academic lecturing process by concentrating on lecturers' didactic behaviours, and other techniques that can be used to visualise the lecture topic. It is stipulated that this can be achieved by automating, summarising, and providing quick feedback to the teachers by using machine learning (ML) methods applied to the video recordings of their lectures.

The overarching research question is: is it possible to apply already existing ML-based methods to design a system that will have as its input a recording of a lecture, and as an output a set of recommendations for the teacher.
To achieve this goal, the didactic features found in the lecture recordings are annotated, and machine learning models are trained to recognise them. Detailed experiments, dedicated to the selected ML models, are available in~\cite{ouraied}. They have been used to set up the proposed approach, which can be summarized as follows:
\begin{enumerate}
\item establishing a set of didactic features feasible to annotate and detect automatically in lecture videos,
\item collecting a dataset of lecture video recordings (that contain the selected features),
\item designing a set of deep learning models trained to determine the presence of the selected didactic feature in the collected dataset -- the video recordings. 
\end{enumerate}

This contribution more precisely describes the captured set of didactic features, and relates them to the teaching practices. Moreover, the reported research has been focused on designing and implementing a prototype, supported with ML models, to automatically recognise the selected didactic features and to prepare summaries and visualisation for the lecturers. Another functionality of the implemented prototype is the possibility of collecting additional datasets and annotating them automatically. This, in turn, allows the proposed system to be further extended and fine-tuned. 

The remainder of this work is structured as follows. 
Section~\ref{sec:dataset} describes the recognized didactic features and the collected dataset. Next, Sections~\ref{sec:design}, \ref{sec:ML-module}, and~\ref{sec:results}  introduce the system prototype and the results, including the interface for the feature visualisation. Finally, the work is summarized in~Section~\ref{sec:conclusions}. There the main findings, and limitations to be considered in the future research, are presented.

\section{The Didactic Features and the Collected Dataset}\label{sec:dataset}


Let us start the presentation by discussing the \textit{didactic features} that have been selected and the reasons for their selection.

\subsection{Didactic Features Captured in the Research}
At first, the teaching practices have been thoroughly analysed, following the approaches described in~\cite{Piburn00reformedteaching,web-stp}. This allowed to establish which features can be used as standards for, objectively, assessing the lessons. 
Next, a set of didactic features, particularly valuable from the teaching point of view have been selected, and adapted to the academic lecturing use case, considered in this work.

\begin{figure}[htpb]
    \centering
    \includegraphics[width=\textwidth]{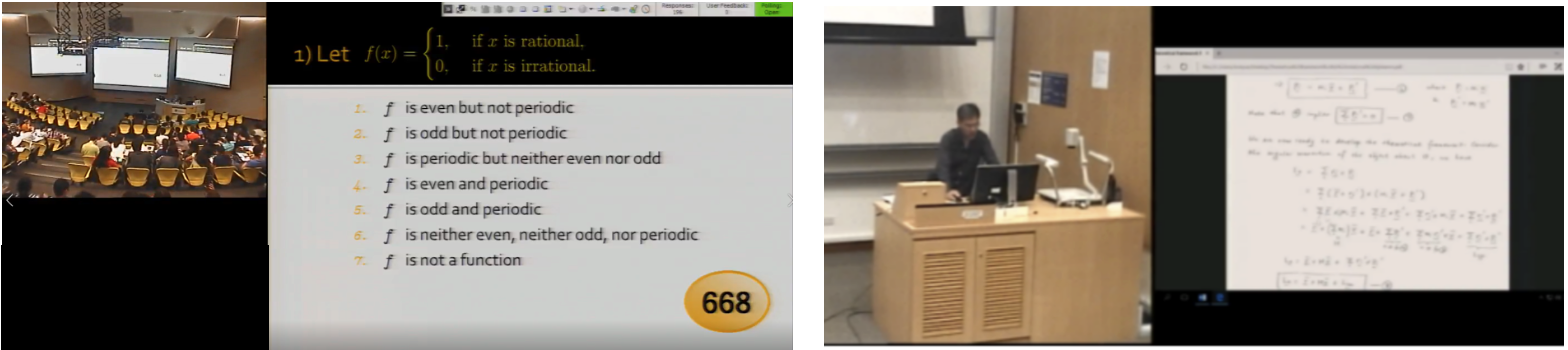}
    \caption{Example frames from the recorded lectures in the collected dataset; this dataset has been also utilized in~\cite{ouraied}}
    \label{fig:example_video}
\end{figure}

Obviously, it has to be possible to detect the selected didactic features, using the appropriately selected and adapted ML models, applied to the lecture video recordings. Figure~\ref{fig:example_video} presents an example of frames that were extracted from lecture recordings (found in the collected dataset). Note that layouts of such frames often consist of two regions -- one originates from the camera view focused on the lecturer, while the second one is a direct input from a computer screen (typically, presentation slides). 

The results of analysis of pertinent literature have been summarized in Table~\ref{tab:features}. Here the selected features are presented and their correspondence to the main categories of Singapore Teaching Practices taxonomy~\cite{web-stp}, e.g., ''activating prior knowledge,'' ''arousing interest'', indicated. In particular the set of actions, performed by the lecturer, such as ''giving class outline'' or ''writing on a whiteboard'' has been selected. In what follows, these teachers' actions are also called the didactic features. Additionally, the  slides' features that help understand the lecture, have been collected.
Note that the qualitative and quantitative didactic features comprise (a) slide characteristics, and (b) different educational behaviours of teachers explaining scientific concepts, e.g. ''plots on slides'', ''asking questions to the students'', or ''actively explaining on the board''. 
Taking into account the goal of capturing didactic features using ML-based approaches, in the Table, the selected features have been divided into three categories: (1) the ones that can be automatically detected from a video stream, i.e. images or videos; (2) those that need to be detected only from the lecture audio stream; and (3) those features that can be detected based on both sources. 
\begin{table}[htpb]
\centering
    \begin{tabular}{p{0.4\textwidth}p{0.6\textwidth}}
\hline
\textbf{Feature} & 
\textbf{Singapore Teaching Practices areas}\\  \hline
\multicolumn{2}{c}{Audio-based features} \\\hline\hline
Asking questions  
& Activating prior knowledge; Encouraging learner engagement; Using questions to deepen learning \\ 
Giving questions to students: rhetorical, comprehension questions  
& Using questions to deepen learning; Arousing interest; Providing clear explanation \\ 
Students are asking questions and generating their ideas     
& Facilitating collaborative learning; Empowering learners; Encouraging learner engagement \\ 
Laughter 
& Arousing interest  \\ 
Discipline 
& Maintaining positive discipline \\ 
Students' discussion 
& Facilitating collaborative learning; Empowering learners; Encouraging learner engagement \\ 
Use of voice intonation to emphasize more important issues/topics 
& Encouraging learner engagement; Pacing and maintaining momentum \\ 

\hline
\multicolumn{2}{c}{Visual features (that can be detected on slides or the view of the teacher)}\\\hline\hline
Films or animations in slides 
& Providing clear explanation; Arousing interest  \\ 
Images in slides  
& Providing clear explanation; Arousing interest  \\ 
Charts in slides 
& Providing clear explanation; Arousing interest \\ 
Showing websites  
& Providing clear explanation; Arousing interest \\ 
An active teacher stands by slides and explains them  
& Encouraging learner engagement \\ 
Movement across podium 
& Encouraging learner engagement  \\ 
Writing on a whiteboard  
& Providing clear explanation; Arousing interest  \\ 
Writing on slides 
& Providing clear explanation; Arousing interest \\ 
Demonstration 
& Arousing interest; Providing clear explanation\\ 
Eye contact 
& Encouraging learner engagement \\ 
\hline
\multicolumn{2}{c}{Visual and/or audio-based features}\\\hline\hline
Referring to the bibliography, other research  
& Deciding on teaching aids and learning resources \\ 
Session on tests 
& Activating prior knowledge; Checking for understanding and providing feedback; Assessment  \\ 
Giving hints on how to do something 
& Activating prior knowledge; Encouraging learner engagements  \\ 
Organization: giving class outline, clearly indicating a transition from one topic to another 
& Setting expectations and routines; Establishing interaction and rapport, determining lesson objectives; Deciding on teaching aids and learning resources; Pacing and maintaining momentum \\ 
Assignments 
& Supporting self-directed learning; Setting meaningful assignments \\ 
Summing up 
& Concluding the lesson  \\ 
\hline
\end{tabular}
\caption{Set of didactic features implemented in this and the previous work~\cite{ouraied}}
\label{tab:features}
\end{table}

\subsection{Collecting Dataset}
\begin{figure}[!ht]
    \includegraphics[width=\textwidth]{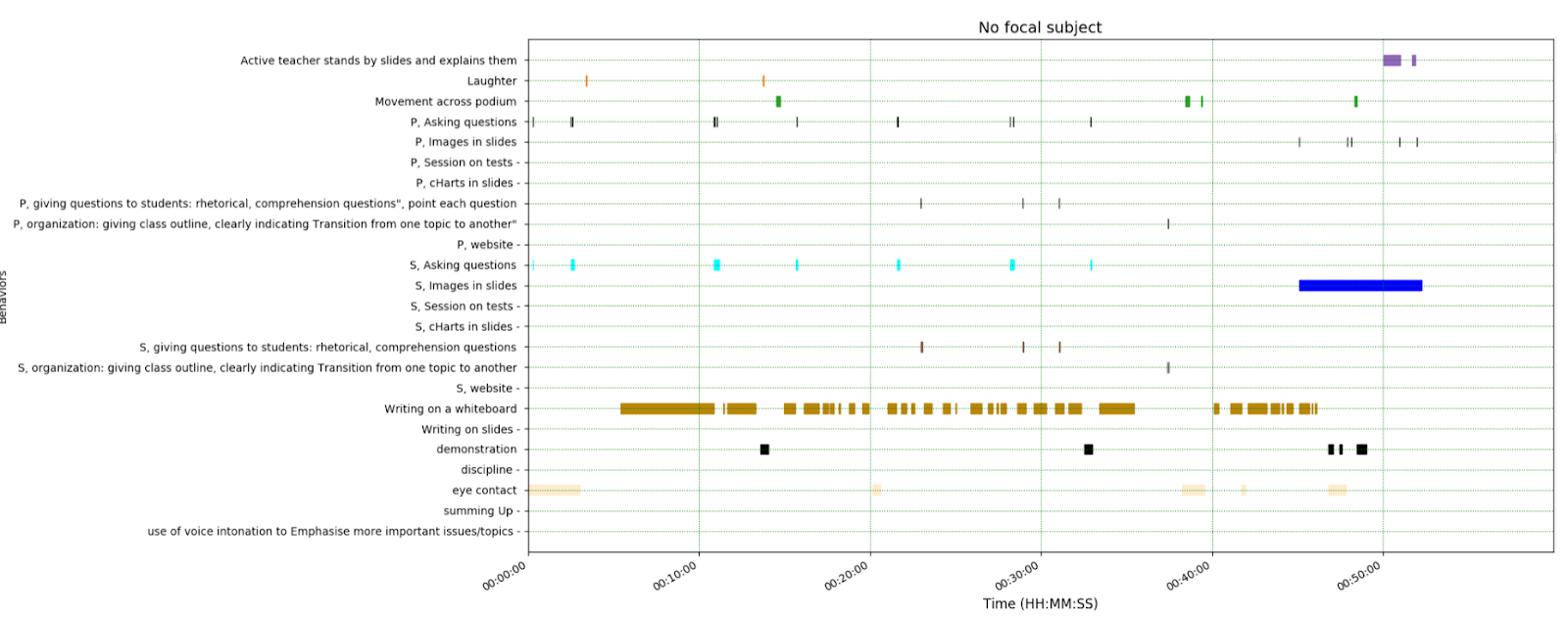}
    \caption{Example annotations in the dataset, a view from BORIS; P -- indicated point event, S -- state event}
    \label{fig:example_annotation}
\end{figure}

After conceptualizing the didactic features, a dataset with those features has been collected and annotated. 
Preparing the annotation dataset is a challenging, complex, and time-consuming task. 
In the case reported here, the process started with designing a set of clear and understandable features (as described above), continued through annotator training and coding. Next, the annotation cleaning and verification ensued, and the process finished with selecting the gold standard for training and testing ML models. This required the combined efforts of multiple people and the application of programs and automation tools. In the annotation process, to facilitate annotations of life events (the features) occurring through time, BORIS -- a free and open-source  Behavioral Observation Research Interactive Software (BORIS is dedicated for video/audio coding and live observations)~\cite{BORIS} has been used
. 
(see, Figure~\ref{fig:example_annotation}).

Finally, the partner university (Nanyang Technological University in Singapore) provided 128 lectures related to the Physics discipline (delivered by multiple lecturers), recorded in English (on average, one lecture lasts an hour and a half).\footnote{We have an ethical acceptance from the NTU  to use this dataset in the reported experiments.} At least three independent research assistants annotated every lecture. Each individual data element is a full-length annotation of didactic features present in a given lectures. Cumulatively, 380 observations were collected.
Each feature was identified as either a state or point event. State events last a specific time, i.e., they have intervals with a precisely defined start time and end time. Point events occur at a given moment and show the changes in the behaviours or the occurrence of a feature, without its specific anchoring in a given time interval. An event means a single didactic feature in a given lecture, annotated by a research assistant. An observation is a set of events for one lecture annotated by one research assistant. An annotation is a set of all observations created for one lecture.
The annotations were then used to train machine learning models.

In the next project phase, various methods how to pre-process and prepare the deep learning models to detect the selected features and how to design ML models to cope with them have been tried. It was necessary to prepare methods that allowed to split two sources of information within the recordings, i.e. the slide and the teacher views. Afterwards, both or separate views have been used to design ML models to apply to the visual features. Moreover, ML techniques needed to identify features in the the audio streams have been selected. Moreover, transcription models have been applied to the collect texts -- the audio stream of the lectures. Detailed results of tests run with all explored techniques have been reported in~\cite{ouraied}.  Here, the models have been improved with the latest state-of-the-art techniques, which have been utilize to build the system prototype.
The overarching goal of the process was to design and implement an interface that can automatically help teachers to get feedback and to visualize their lectures so that they can derive ways to improve their performance. 

\section{System Design}\label{sec:design}
In the proposed system design, it has been assumed that the automatically detected features will be used for the objective summary evaluation, to provide feedback for the academic teachers on their lecturing style and effectiveness.
To achieve this goal the prototype's graphic interface allows to upload a lecture video and get a list of didactic features occurring within it. 
Overall, the developed application is organized in a way that does not require technical knowledge. The user should obtain the annotation of didactic features, view the visualisation of features detected by the model, and see which part of the lecture was automatically annotated. In addition, users should be able to see the visualisation of trends, based on all of the lectures uploaded and annotated in the previous sessions. 

\begin{figure}[htpb]
    \centering
    \includegraphics[width=\linewidth]{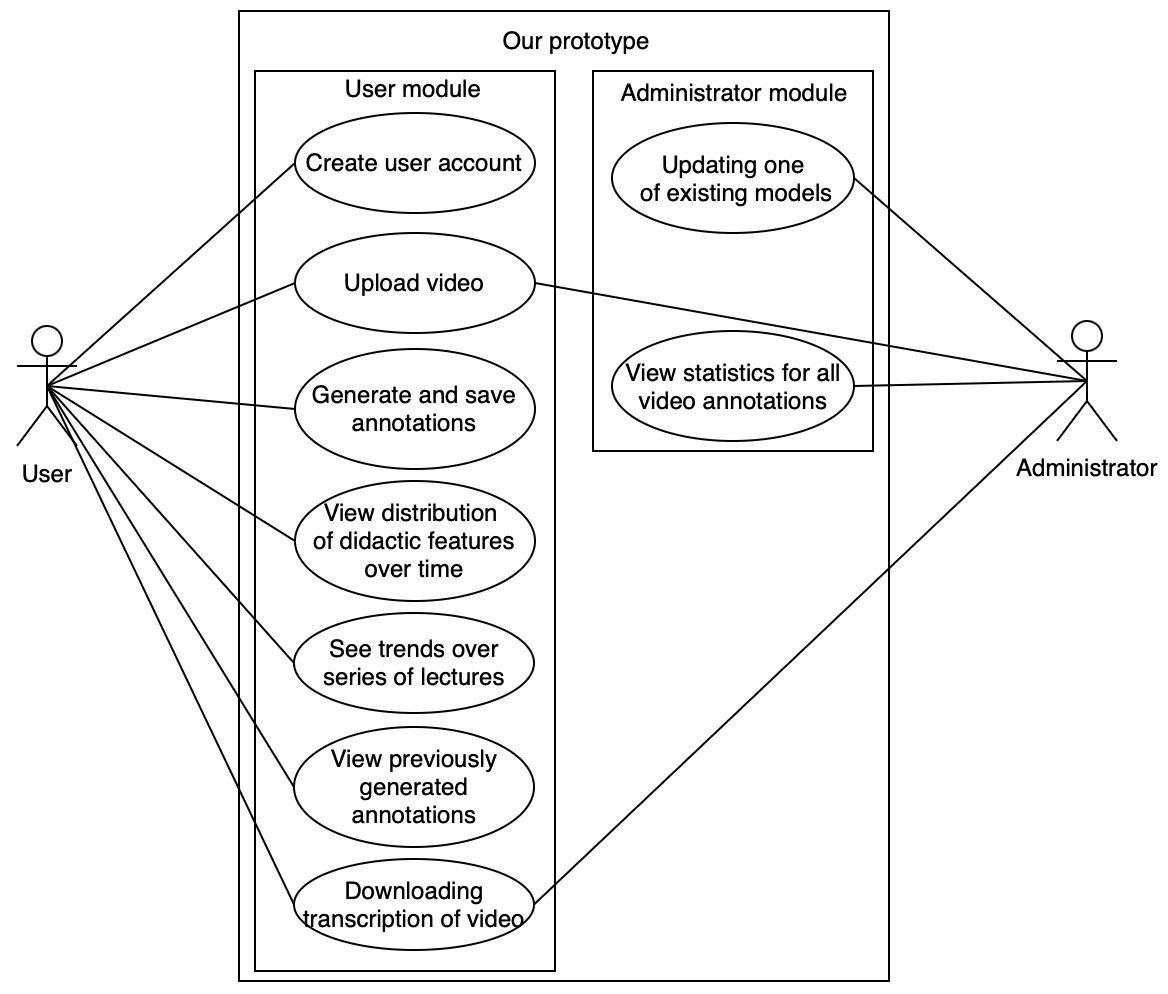}
    \caption{System use case diagram}
    \label{fig:usecases}
\end{figure}

The prototype's functional requirements were defined assuming that the developed application has two types of users: Lecturer (User) and Administrator. Without the need for ML knowledge, the lecturer uses the application to get feedback on their lecturers (see, Figure~\ref{fig:usecases}). The Administrator, on the other hand, has machine learning knowledge and can change or upgrade the models, available in the system.

Moreover, the system was designed to fulfil the standard non-functional requirements for web applications, i.e. (i) usability -- intuitive interface, widely-used programming standards to reuse the application source code, reliability; (ii) performance -- short time of response for ML-models; and (iii) supportability -- easy-to-change ML models, easy extending source code with up-to-date standards (e.g. Python, dash, plotly, pytorch, whisper, sqlite3). 

\begin{figure}[htpb]
\centering
\includegraphics[width=.7\textwidth]{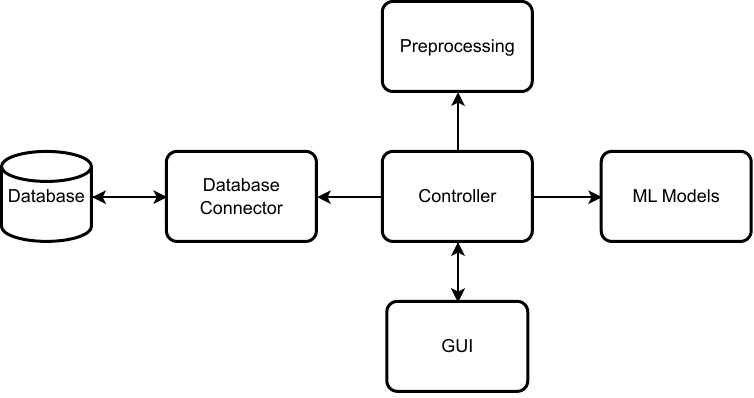}
\caption{Implemented prototype's design -- module diagram}
\label{module_diagram}
\end{figure}

The implemented prototype comprises six modules: \textit{Controller, GUI, Preprocessing, ML Models, Database Connector} and \textit{Database} (see, Figure~\ref{module_diagram}). The \textit{Controller} module communicates with the other modules, and runs the application. The \textit{Preprocessing} module transforms video into a format that can be passed to the models that are being trained, e.g. a sequence of images, only audio stream. Preprocessing of the video consists of two steps: (A) the transcription is created with the \texttt{openai-whisper} package; (B) the JSON file with generated transcription is transformed into a data frame with the start and the end timestamps of the sentence. The \textit{ML Models} module stores imported from pytorch binary files, with the model weights and other parameters, which are then used to create predictions.
The \textit{GUI} module is responsible for creating the graphic interface and handling input from the user. To make the interface, the \texttt{dash} framework,\footnote{\url{https://dash.plotly.com}} was used, as it allows integration of interactive plots into the web application. The graphic interface allows the users to upload a new video, get transcription, and view the analysis of predicted didactic features. The \textit{Database} module is where all datasets (including videos uploaded by the users) are stored.

\section{Machine Learning Module}\label{sec:ML-module}

In the developed prototype, the audio was used to create the transcription and, in subsequent steps, the transcription was used for building text-based models. Let us now describe it in some more detail.

\subsection{Transcription Process}

Even though the errors of the transcription models could be propagated to the classification model, the method based on transcription achieved generally better results than the method employing audio characteristics, and this transcription and training, or predicting, pipeline takes less computation power. The transcription results were tested on six short, manually transcribed lecture fragments. Ultimately, selected was the tool with the best word error rate (WER) metric results~\cite{mccowan2004use} -- see Table~\ref{tab:transcription}.
\begin{table}[htpb]
\centering
\begin{tabular}{lcccccccc}
~~ & \textbf{Deepspeech~~} & \textbf{Azure~~} & \textbf{Watson~~} & \textbf{Vosk~~} & \textbf{A-S-R~~} & \textbf{Jasper~~} & \textbf{GCP~~} & \textbf{Whisper~~}\\
\hline
~~  & 50.18 & 19.02 & 47.15 & 43.62 & 63.04 & 49.97 & 34.26 & \textbf{16.81}\\
\end{tabular}
\caption{WER on test dataset for lecture transcription}
\label{tab:transcription}
\end{table}

\subsection{Text-based Models}
The transcriptions obtained from the existing transcriptions tools were used to train two machine learning models: for the binary classification task, noting whether the given sentence contains a question, and for the multiclass classification task, which marks all relevant text-based didactic features. 

The state-of-the-art transfer learning with Transformer BERT-based models provided by HuggingFace~\cite{liu2020roberta}\footnote{https://huggingface.co/{roberta-base}}, was used.
The dataset classification tasks are selected as the downstream tasks each time the training procedure and the pretraining are performed on another, larger dataset, the BookCorpus dataset~\cite{book_corpus}\footnote{https://huggingface.co/datasets/bookcorpus}. A feed-forward neural network head is added at the end of the Transformers, and predicts classes from the dataset. Default uncased BERT-Base model hyperparameters were not changed during the hyperparameter optimization process, and instead, the focus was on fine-tuning the last layer, the feed-forward network. 

The models used for the recognition of features regarding questions achieved a precision of about 20\%, recall of about  50\%, and F1 score of about 30\%. It is worth mentioning that the model recognised six features at once, so the random results would be about 16\%. Hence, the obtained results seem reasonable, mainly because the dataset was relatively small, and the task was challenging to annotate.

\section{System Interface}\label{sec:results}

\begin{figure}[htpb]
        \centering
            \includegraphics[width=\textwidth]{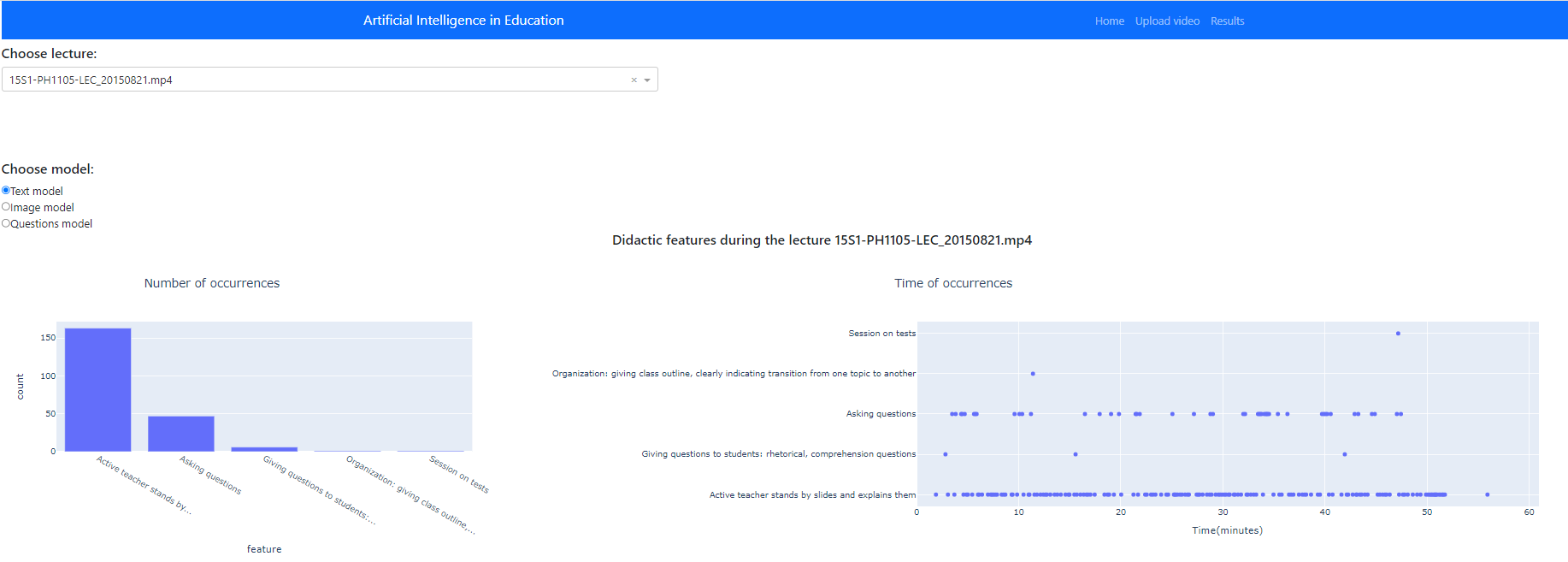}
        \caption{Displaying plots with a summary of didactic features that appeared during the selected lecture: number and time of occurrences of features predicted by the ML model}
        \label{fig:plots}
\end{figure}

The implemented prototype automatically produced summaries of the detected didactic features. Figure~\ref{fig:plots} shows an interface with the results for a single lecture recording with text-based features, e.g. 'asking questions' and 'organizational issues'. This visualisation contains a number of occurrences of each feature during a given lecture and varying time spans for each feature. The first plot is a bar graph with the number of occurrences of every feature. The second plot's x-axis shows time in minutes; on the y-axis, there are didactic features. Every point represents a single occurrence of a feature in the lecture. 

\begin{figure}[htpb]
        \centering
            \includegraphics[width=\textwidth]{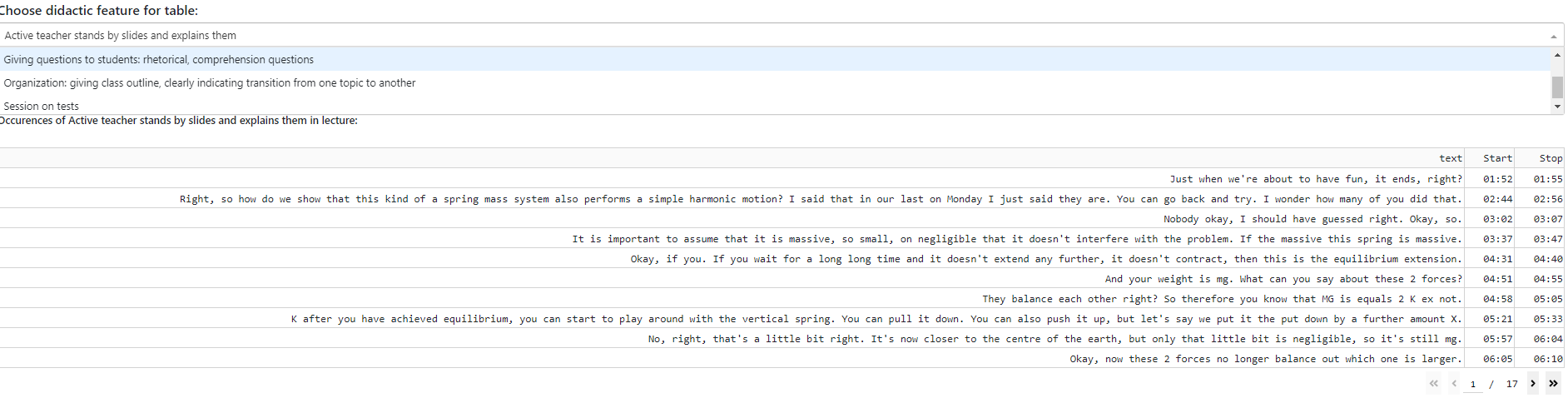}
        \caption{Displaying table with text fragments connected to chosen didactic feature and time interval of its occurrence.}
        \label{fig:plot-trasncription}
    \end{figure}

The user can also see and download the transcription results combined with the results of the text-based models, showing in which sentences any text-based features occurred. This functionality is available when the Text or the Questions model is selected. A dropdown list with features that have been recognized in the selected lecture(s) is displayed. After selecting one, the user can see a table with all moments where the model predicted the occurrence of the feature, with text, time of start, and time of the end of each occurrence. An example of a table look is shown in Figure~\ref{fig:plot-trasncription}.
      
\section{Concluding Remarks}\label{sec:conclusions}

The goal of the reported research was to develop and implement a prototype of an academic lecturer support system, based on application of ML-models to the lecture video recordings.
To achieve this goal a set of didactic features was selected and adapted to objectively and quantitatively evaluate and compare academic lectures. Moreover, the implemented prototype supports automating the annotation process of lecture video recordings using pre-trained machine-learning models. Utilizing these annotations, summaries and visualizations of the didactic features observed within the lectures have been created. The developed system is anchored in in-depth analysis of required system's functionalities, which guided the design of its architecture, dataflow, and interfaces. 
The modularity of the designed system allows easy addition of ne ML models, if such models will be found to be promising for the task at hand.
Obtained results, validating the approach, are reasonable, taking into account the size of the dataset used for model training.



The primary target of future research has to be the collection of a larger dataset to be used for model training. Here, what is worthy repeating is that the developed prototype can be used to form a partially-supervised continual learning loop. Specifically, the existing system can be used for preliminary didactic feature extraction, helping annotate future videos (human-in-the loop). These videos can then be used to up-train the model(s). 

The dataset and the system's software is available upon request from the corresponding author. However, note that restrictions related to the GDPR (and other pertinent regulations) apply, and applicable documents will guide granting access to both the dataset and the code.

\section*{Acknowledgments}
This work is done in cooperation with Nanyang Technological University, in the frame of OMINO (Overcoming Multilevel Information Overload) grant (no 101086321) funded by the European Union under the Horizon Europe and by the Polish Ministry of Education and Science  within the International Projects Co-Financed program.
(However, the views and opinions expressed are those of the authors only and do not necessarily reflect those of the European Union or the European Research Executive Agency. Neither the European Union nor the European Research Executive Agency can be held responsible for them.)

This research was also carried out with the support of the Faculty of Mathematics and Information Science at Warsaw University of Technology, its Laboratory of Bioinformatics and Computational Genomics, and the High-Performance Computing Center.

\bibliographystyle{unsrt}  
\bibliography{references}

\begin{thebibliography}{10}

\bibitem{kuvcak2018machine}
Danijel Kucak, Vedran Juricic, and Goran Dambic.
\newblock Machine learning in education - a survey of current research trends.
\newblock In {\em DAAM Int. Symp. on Intelligent Manufacturing and Automation}, 2018.

\bibitem{unescogenai}
Fengchun Miao and Wayne Holmes.
\newblock {\em Guidance for generative AI in education and research}.
\newblock UNESCO, 2023.

\bibitem{anand2018recursive}
VK~Anand, SK~Abdul Rahiman, E~Ben George, and AS~Huda.
\newblock Recursive clustering technique for students' performance evaluation in programming courses.
\newblock In {\em IEEE Majan Int. Conf.}, 2018.

\bibitem{zhu2015machine}
Xiaojin Zhu.
\newblock Machine teaching: An inverse problem to machine learning and an approach toward optimal education.
\newblock In {\em AAAI Conference on AI}, 2015.

\bibitem{holmes2023artificial}
Wayne Holmes, Sun Meng, and Li~Yuan.
\newblock Artificial intelligence and education: Digging beneath the surface.
\newblock {\em The Chinese Journal of ICT in Education}, 2023(2):16--26, 2023.

\bibitem{segal2017keeping}
Avi Segal, Shaked Hindi, Naomi Prusak, Osama Swidan, Adva Livni, Alik Palatnic, Baruch Schwarz, and Ya’akov Gal.
\newblock Keeping the teacher in the loop: Technologies for monitoring group learning in real-time.
\newblock In {\em Artificial Intelligence in Education: 18th International Conference, AIED 2017, Wuhan, China, June 28--July 1, 2017, Proceedings 18}, pages 64--76. Springer, 2017.

\bibitem{broussard2021interface}
David~M Broussard, Yitoshee Rahman, Arun~K Kulshreshth, and Christoph~W Borst.
\newblock An interface for enhanced teacher awareness of student actions and attention in a vr classroom.
\newblock In {\em 2021 IEEE Conference on Virtual Reality and 3D User Interfaces Abstracts and Workshops (VRW)}, pages 284--290. IEEE, 2021.

\bibitem{tarantini2022reflective}
Eric Tarantini.
\newblock Reflective teacher education in the digital age: 360° video reflection and ai-based developments.
\newblock In {\em Open and Inclusive Educational Practice in the Digital World}, pages 213--231. Springer, 2022.

\bibitem{AIcoach}
Lauraine Langreo.
\newblock Can {AI} do teacher observations.

\bibitem{guo2021ai}
Junqi Guo, Ludi Bai, Zehui Yu, Ziyun Zhao, and Boxin Wan.
\newblock An ai-application-oriented in-class teaching evaluation model by using statistical modeling and ensemble learning.
\newblock {\em Sensors}, 21(1):241, 2021.

\bibitem{ouraied}
Anna Wr{\'o}blewska, J{\'o}zef Jasek, Bogdan Jastrzebski, Stanis{\l}aw Pawlak, Anna Grzywacz, Siew~Ann Cheong, Seng~Chee Tan, Tomasz Trzci{\'{n}}ski, and Janusz Ho{\l}yst.
\newblock Deep learning for automatic detection of qualitative features of lecturing.
\newblock In Maria~Mercedes Rodrigo, Noburu Matsuda, Alexandra~I. Cristea, and Vania Dimitrova, editors, {\em Artificial Intelligence in Education}, pages 698--703, Cham, 2022. Springer International Publishing.

\bibitem{Piburn00reformedteaching}
Michael Piburn and Daiyo Sawada.
\newblock Reformed teaching observation protocol {(RTOP)} reference manual.
\newblock Technical report, Arizona Collaborative for Excellence in the Preparation of Teachers (ERIC ED447205), 2000.

\bibitem{web-stp}
{Singapore Ministry of Education}.
\newblock {The Singapore Teaching Practice} {(STP)}, 2018.

\bibitem{BORIS}
Olivier Friard and Marco Gamba.
\newblock Boris: a free, versatile open-source event-logging software for video/audio coding and live observations.
\newblock {\em Methods in Ecology and Evolution}, 2016.

\bibitem{mccowan2004use}
Iain~A McCowan, Darren Moore, John Dines, Daniel Gatica-Perez, Mike Flynn, Pierre Wellner, and Herv{\'e} Bourlard.
\newblock On the use of information retrieval measures for speech recognition evaluation.
\newblock {\em Reporte de investigación 04-73}, 2005.

\bibitem{liu2020roberta}
Yinhan Liu, Myle Ott, Naman Goyal, Jingfei Du, Mandar Joshi, Danqi Chen, Omer Levy, Mike Lewis, Luke Zettlemoyer, and Veselin Stoyanov.
\newblock Ro{\{}bert{\}}a: A robustly optimized {\{}bert{\}} pretraining approach, 2020.

\bibitem{book_corpus}
Yukun Zhu, Ryan Kiros, Rich Zemel, Ruslan Salakhutdinov, Raquel Urtasun, Antonio Torralba, and Sanja Fidler.
\newblock Aligning books and movies: Towards story-like visual explanations by watching movies and reading books.
\newblock In {\em IEEE ICCV}, 2015.

\end{thebibliography}

\end{document}